\documentclass{article} % For LaTeX2e
\usepackage{iclr2018_workshop,times}
\usepackage{hyperref}
\usepackage{url}
\usepackage{graphicx}
\usepackage{multirow}
\usepackage{wrapfig}
\usepackage{amsmath}
\usepackage{caption}
\usepackage{float}
\usepackage{subcaption}
\usepackage[compact]{titlesec}
\usepackage{siunitx}
\usepackage[percent]{overpic}
\usepackage{tikz}

\title{IamNN: Iterative and Adaptive Mobile Neural
Network for efficient image classification}

\author{Sam Leroux\thanks{The research work was done during an internship at NVIDIA research, Santa Clara, California, USA} , Pieter Simoens \& Bart Dhoedt  \\
Ghent University - imec, IDLab\\
Department of Information Technology\\
Technologiepark-Zwijnaarde 15\\
B-9052 Ghent, Belgium\\
\And
Pavlo Molchanov, Thomas Breuel \& Jan Kautz \\
NVIDIA\\
2788 San Tomas Expressway\\
Santa Clara, California\\
USA\\
}

\begin{document}

\maketitle
%\vskip -10pt
\begin{abstract}
Deep residual networks %with skip connections
(ResNets) made a recent breakthrough in deep learning. The core idea of ResNets is to have shortcut connections between layers that allow the network to be much deeper while still being easy to optimize avoiding vanishing gradients. These shortcut connections have interesting side-effects that make ResNets behave differently from other typical network architectures. In this work we use these properties to design a network based on a ResNet but with parameter sharing and with adaptive computation time. The resulting network is much smaller than the original network and can adapt the computational cost to the complexity of the input image.
%We present results on different benchmark datasets and show that we can reduce the model size by 90\% and the computational cost by half at a modest decrease in classification accuracy.
\end{abstract}

\section{Introduction and related work}
After their impressive results on the ILSVRC2015 challenge, deep residual networks% with skip connections (ResNets)
~\citep{He_2016_CVPR} quickly became one of the default architectures for computer vision tasks. % such as image classification and object detection. %Different breakthroughs such as better nonlinearities \citet{nair2010rectified}, better initialization methods \citep{glorot2010understanding}, \citep{he2015delving}, \citep{xie2017all} and normalization of the intermediate activations (\citep{ioffe2015batch}) allowed us to build increasingly deep networks, yet very deep networks were still hard to optimize.
%ResNets solved this by introducing a different flow.
Instead of just
stacking layers on top of each other where each layer has to transform the output of the previous layer ($h_{i+1} = F_i(h_i)$), they add \textit{skip connections}, identity mappings that copy the input of the layer to the output. The layer then learns a \textit{residual} to add to the input ($h_{i+1} =  h_i + F_i(h_i)$)
. This makes it easier to optimize the network and allows us to build networks with hundreds or even thousands of layers. ResNets are closely related to Highway networks \citep{srivastava2015highway} where the flow of information is regulated by gates instead of using a fixed skip connection.

The residual connections have some very interesting properties in addition to increasing the maximum depth of the network. Various works have shown that ResNets are remarkably robust against deleting or reordering layers from a trained network \citep{veit2016residual} \citep{srivastava2015highway} while this behavior is not found in the more traditional architectures. \citet{veit2016residual} argue that ResNets should be interpreted as an exponential ensemble of shallower models. Because of the shortcut connections there is a large number of possible paths with different lengths and although all these paths are trained simultaneously they do not strongly depend on each other. Deleting some layers of the network corrupts certain paths but since there is an exponentially large number of paths 
the overall effect on the final accuracy is limited. \citet{greff2016highway} on the other hand argue that the layers in a ResNet do not learn completely new representations but instead they 
gradually refine the features extracted by the previous layers. In this \textit{unrolled iterative estimation view} successive layers cooperate to compute a single level of representation. \citet{jastrzebski2017residual} provided a formal view of iterative feature refinement and showed that a each residual layer refines the feature representation to reduce the loss with respect to the hidden representation.

In this work we follow the iterative estimation view and exploit these properties to design networks with less parameters and a smaller computational cost. To reduce the model size we enforce a strict parameter sharing between the layers in the network. Since each layer refines the features extracted by the previous layer instead of learning completely new features, it seems plausible that there is a certain redundancy in having different parameters for each layer. Sharing parameters between layers has been proposed before \citep{jastrzebski2017residual} \citep{boulch2017shaResNet} and both papers report impressive model size reductions.

To reduce the computational cost we incorporate an Adaptive Computation Time (ACT) mechanism. In a traditional ResNet every sample follows the exact same path through the network where every layer refines the features from the previous layer. Not all samples are equally hard to classify and some of them will need more refinement steps than others. With ACT we can include a component that will decide how many steps are needed for each given input sample. ACT was introduced by \citet{graves2016adaptive} for recurrent neural networks where the network adapts the number of calculation steps to the complexity of the input. ACT can also be applied to networks for image recognition.
\citet{figurnov2016spatially} introduced a ResNet based architecture where ACT was used to decide to evaluate or to skip certain layers. They even further improved this to Spatial ACT (SACT) which adapts the amount of computation between spatial positions.

We combined these two ideas of parameter sharing and adaptive computation time to design a building block for accurate deep neural networks with small model size and adaptive computational cost.
%Combining these 2 ideas of parameter sharing and adaptive computation time requires changes to the original residual block. In this work we propose a new building block for deep networks with a high accuracy but with small model size and computational cost.

\section{Architecture}

\begin{figure}
\vskip -40pt
\begin{minipage}{.5\textwidth}
      \includegraphics[trim={3cm 5cm 5cm 3cm},clip,width=\linewidth]
      {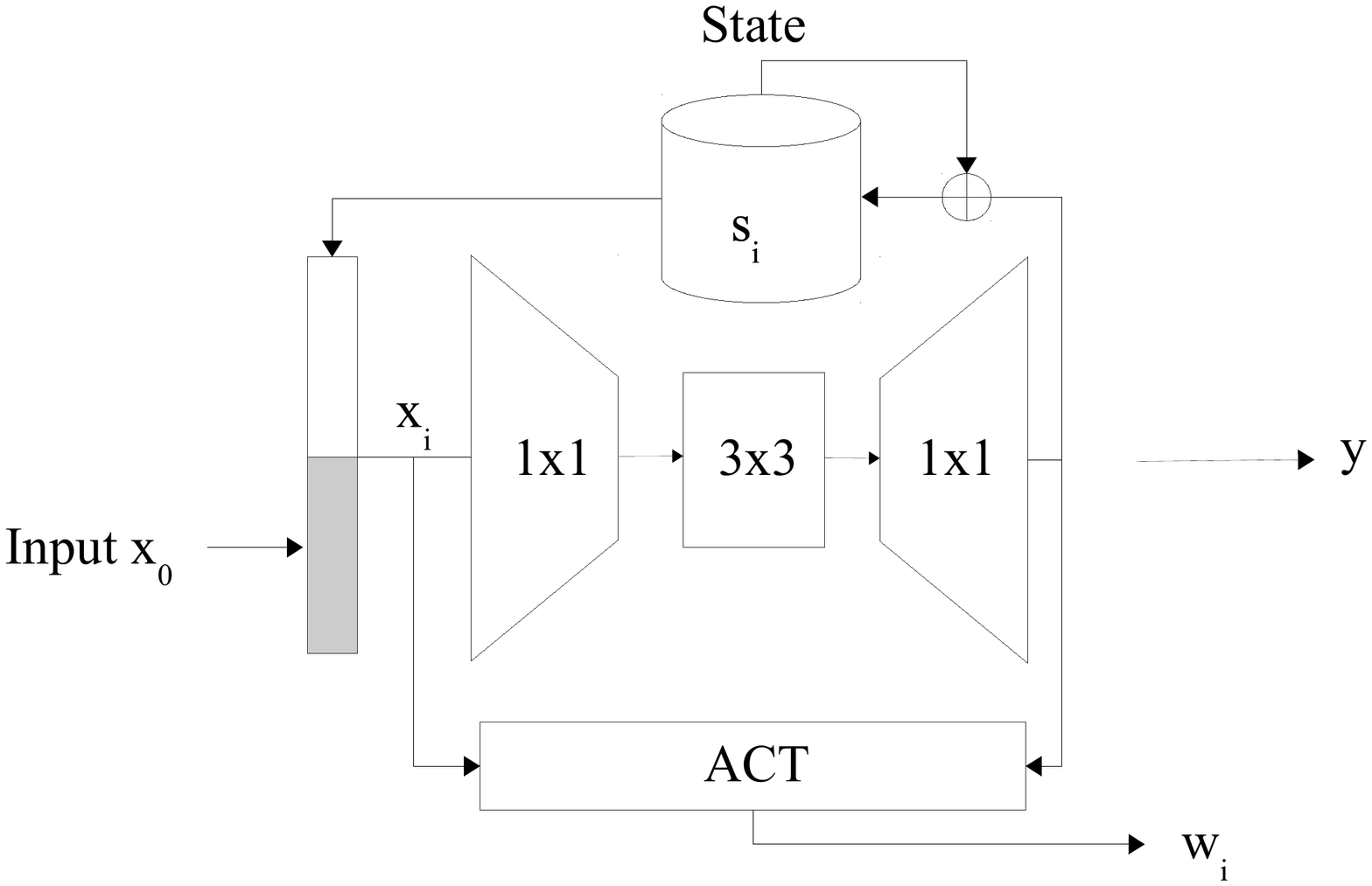}
      %\captionof{figure}{The building block of our network. }
      
\end{minipage}%
\begin{minipage}{.5\textwidth}
	\small
    \begin{align*}
        & x_i = concat(x_0, s_{i-1}) \hspace{23pt} y = x_0 + \sum_{i}^{} w_is_i\\
        & s_i = s_{i-1} + F(x_i) \hspace{37pt} s_0 = 0 \\
		& w_i = ACT(x_i, F(x_i)) \hspace{20pt} \sum_{i}^{} w_i = 1 \\
    \end{align*}
    \vskip -15pt
    \hskip -25pt
    \includegraphics[scale=0.525]{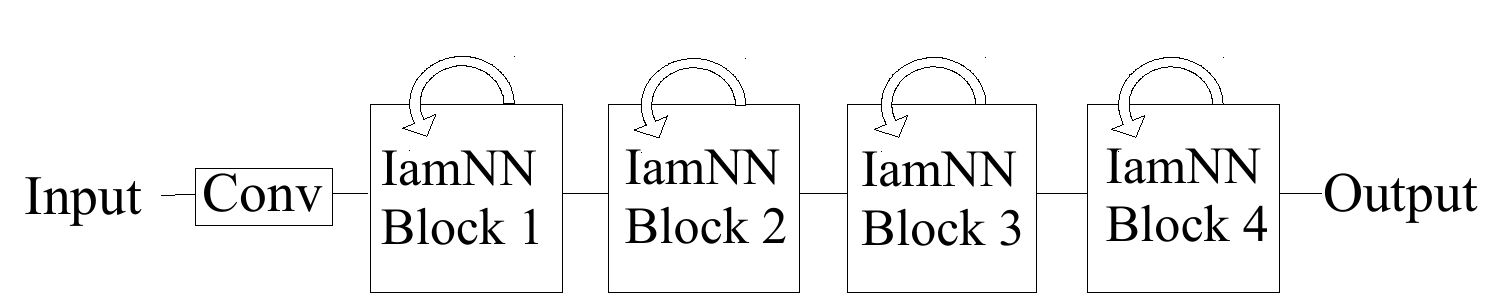}
    %\captionof{figure}{Both a ResNet and the IamNN architecture have four blocks. }
\end{minipage}%
%\caption{Both a ResNet and the IamNN architecture have four blocks (right). We propose to replace the residual blocks with the architecture shown on the left.}
\caption{The IamNN network has the same structure as a ResNet but we replace the many residual units in each block by the architecture on the left which reuses the same weights multiple times.}
\label{fig:architecture}
\end{figure}

A typical ResNet consists of first a convolutional layer with batch normalization and ReLu nonlinearity followed by maxpooling. Then, a sequence of four blocks is stacked where each block consists of multiple stacked residual units. Each residual unit  consists of one or more convolutional layers and a shortcut connection. All residual units in a block operate on the same spatial size. The first convolutional layer in each block uses a convolution with stride 2 to reduce the spatial size.

We propose to replace each block by the architecture shown in figure \ref{fig:architecture} (left). Each residual unit in a block is replaced by an iteration of this module. We use maxpooling between each block to reduce the spatial size. Our block consist of three main parts: a processing block with three convolutional layers, a state buffer where the results are accumulated and the ACT block that decides how many iterations of this block are needed.

The state buffer is used to gradually build the output of the block. The output of each iteration is added to the state which allows for an iterative refinement of the features. The initial state ($s_0$) is initialized with zero values.

The processing block consists of three convolutional layers with a bottleneck structure \citep{He_2016_CVPR}. %Bottleneck structures are common in ResNets because they reduce the computational cost \citep{He_2016_CVPR}. The 1x1 convolutions are used to reduce and to restore the dimension (number of channels) of the representations, leaving the 3x3 convolution to work with smaller input and output dimensions.
At each iteration we concatenate the original input of the block ($x_0$) with the current state ($s_{i-1}$) before passing the concatenated vector through the processing block. The output of the processing block is added to the state. Each convolutional layer is followed by batch normalization, as is common in ResNets. We %however
found that it is not possible to use the same batchnorm statistics for multiple iterations, instead we use a different set of statistics in each iteration. This only incurs a small overhead since the number of batchnorm statistics is very small compared to the full network. The same approach was used in \citet{jastrzebski2017residual} in their experiments with shared weights.

The ACT block is a small two layer fully connected network (2 times 64 hidden units) that decides whether to keep evaluating the block or to move on to the next block.% We first apply global average pooling to the original input of the block, to the current state and to the output of the processing block. %Global average pooling calculates the average for each channel in the input and these average values are given to the ACT network.
We concatenate the current state, the original input and the output of the processing block and apply global average pooling to obtain a vector with a single value for each channel. The ACT network uses this vector to calculate a \textit{halting score} between 0 and 1 (sigmoidal activation).% We keep evaluating the same block as long as the cumulative halting score is smaller than one.
We sum the halting scores of each iteration and as soon as the cumulative halting score reaches one we stop evaluating this block and move on to the next block. We add an additional loss term to the classification loss to encourage the network to use a small number of iterations.% The computational cost of the ACT block is 0.3\% of the total number of operations of the network.

\section{Results}
We trained our models on three benchmark datasets for image recognition: CIFAR10, CIFAR100 and ImageNet. We report the number of parameters, the theoretical number of operations and the test accuracy in table \ref{tbl:results}. We designed our models after a ResNet architecture and constrained them to use at maximum the same number of iterations in each block as the the number of units in that block of the ResNet. ResNet152 for example has four blocks with 3, 8, 36 and 3 residual units respectively. The corresponding IamNN is then constrained to use at maximum 3, 8, 36 and 3 iterations for each block respectively. The computational cost of the IamNN networks varies depending on the complexity of the input image, we report the average required FLOPS over all images in the test set.

For CIFAR10 and CIFAR100 we reduce the number of parameters by 90\% compared to the corresponding ResNet101. The number of FLOPS varies between 0.8 GFLOPs and 2 GFLOPs for both datasets depending on the complexity on the input image. On average we require 1.1 billion and 1.6 billion operations for CIFAR10 and CIFAR100 respectively, a reduction of 56\% and 36\% respectively. For CIFAR10 we obtain a slightly higher accuracy compared to the ResNet, probably because weight sharing acts as a regularizer for the small dataset. On CIFAR100 however the accuracy drops from 79.3\% to 77.8\%.

For the ImageNet dataset we used ResNet152 as a baseline. We are again able to reduce the number of parameters with 90\% and the computations by 65\% the top5 accuracy drops from 93.3\% to 89\%. To illustrate the benefit of iterative refinement using the same weights we also include the result when our IamNN network is only allowed one iteration per block. In this case there is no iterative refinement possible and the network obtains a top5 accuracy of 83.2\%. We also compare to other architectures that were built to be very efficient in terms of computational cost and memory footprint (MobileNet \citep{howard2017mobilenets}, ShuffleNet \citep{zhang2017shufflenet}, SqueezeNet \citep{iandola2016squeezenet} and GoogleNet \citep{DBLP:journals/corr/SzegedyLJSRAEVR14}). These networks obtain a similar accuracy with a similar number of parameters but at a lower computational cost. This is because we did not change the basic operations of the network (1x1 and 3x3 convolutions). Some of the techniques used in the efficient networks such as depthwise separable convolutions are however orthogonal to our weight sharing and ACT approach and could be used to reduce the computational cost even further.

\begin{figure}
\vskip -40pt
\begin{subfigure}{\textwidth}
  \begin{tikzpicture}
    \node[text width=4cm] at (0,0.4) {easy images};
    \node[text width=4cm] at (12,0.4) {hard images};
  	\draw [->,>=stealth] (-2,0.1) -- (12,0.1);
  \end{tikzpicture}%
\end{subfigure}
\centering
  \begin{subfigure}{\textwidth}
  \centering
    \begin{subfigure}{.1\textwidth}
      \includegraphics[trim={2px 2px 2px 2px},clip, width=\linewidth]{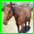}
    \end{subfigure}
    \begin{subfigure}{.1\textwidth}
      \includegraphics[width=\linewidth]{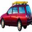}
    \end{subfigure}
    \begin{subfigure}{.1\textwidth}
      \includegraphics[width=\linewidth]{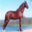}
    \end{subfigure}
    \begin{subfigure}{.1\textwidth}
      \includegraphics[width=\linewidth]{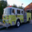}
    \end{subfigure}
    \begin{subfigure}{.1\textwidth}
      \includegraphics[width=\linewidth]{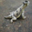}
    \end{subfigure}
    \begin{subfigure}{.1\textwidth}
      \includegraphics[width=\linewidth]{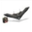}
    \end{subfigure}
    \begin{subfigure}{.1\textwidth}
      \includegraphics[trim={2px 2px 2px 2px},clip,width=\linewidth]{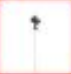}
    \end{subfigure}
    \begin{subfigure}{.1\textwidth}
      \includegraphics[trim={2px 2px 2px 2px},clip,width=\linewidth]{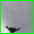}
    \end{subfigure}
    \begin{subfigure}{.1\textwidth}
      \includegraphics[trim={2px 2px 2px 2px},clip,width=\linewidth]{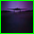}
    \end{subfigure}
    %\caption{CIFAR10}
	%\label{fig:cifar10}
  \end{subfigure}

\begin{subfigure}{\textwidth}
\centering
  \begin{subfigure}{.1\textwidth}
    \includegraphics[trim={2px 2px 2px 2px},clip,width=\linewidth]{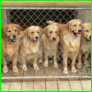}
  \end{subfigure}
  \begin{subfigure}{.1\textwidth}
    \includegraphics[trim={2px 2px 2px 2px},clip,width=\linewidth]{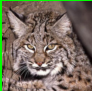}
  \end{subfigure}
  %\begin{subfigure}{.1\textwidth}
  %  \includegraphics[trim={2px 2px 2px 2px},clip,width=\linewidth]{22.png}
  %\end{subfigure}
  \begin{subfigure}{.1\textwidth}
    \includegraphics[trim={2px 2px 2px 2px},clip,width=\linewidth]{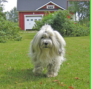}
  \end{subfigure}
  \begin{subfigure}{.1\textwidth}
    \includegraphics[trim={2px 2px 2px 2px},clip,width=\linewidth]{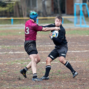}
  \end{subfigure}
  \begin{subfigure}{.1\textwidth}
    \includegraphics[trim={2px 2px 2px 2px},clip,width=\linewidth]{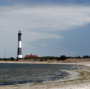}
  \end{subfigure}
  \begin{subfigure}{.1\textwidth}
    \includegraphics[trim={2px 2px 2px 2px},clip,width=\linewidth]{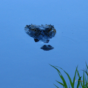}
  \end{subfigure}
  \begin{subfigure}{.1\textwidth}
    \includegraphics[trim={2px 2px 2px 2px},clip,width=\linewidth]{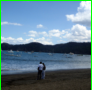}
  \end{subfigure}
  \begin{subfigure}{.1\textwidth}
    \includegraphics[trim={2px 2px 2px 2px},clip,width=\linewidth]{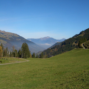}
  \end{subfigure}
  \begin{subfigure}{.1\textwidth}
    \includegraphics[trim={2px 2px 2px 2px},clip,width=\linewidth]{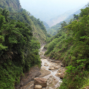}
  \end{subfigure}
  %\caption{ImageNet}
  %\label{fig:ImageNet}
  \end{subfigure}
\caption{The spectrum of easy (left) to hard (right) to classify images for the network trained on CIFAR10 (top row) and ImageNet (bottom row). The network automatically adapts the computational cost to the complexity of the input image.}
\label{fig:test}
\end{figure}
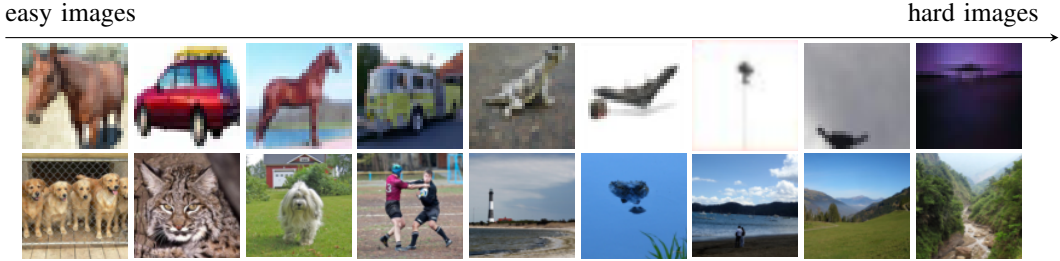

Figure \ref{fig:iterations} shows how many iterations are used for each block in the ImageNet network. The vertical line indicates the number of residual units in the corresponding ResNet (= the maximum number of iterations for the IamNN). Figure \ref{fig:test} illustrates how the network is able to adapt the computational cost to the complexity of the image. We show some typical images sorted from fast (left) to slow (right) to classify. The computation cost varies between 0.7G and 2G FLOPS for CIFAR10 and between 2.5G and 9G FLOPS for ImageNet.

% source results:
% cifar10 ResNet101 accuracy: https://github.com/kuangliu/pytorch-cifar
% cifar100 ResNet101 accuracy: Own result
% ResNet18 accuracy: https://github.com/jcjohnson/cnn-benchmarks
% ResNet152 accuracy: https://github.com/KaimingHe/deep-residual-networks
% ResNet18/ ResNet152 flops: Deep Residual Learning for Image Recognition paper
% ResNet152 params: https://arxiv.org/pdf/1701.04923
% ResNet18 params: 
% ShaResNet accuracy and params: ShaResNet: reducing residual network parameter number by sharing weights
% Mobilenet accuracy flops and params: https://github.com/marvis/pytorch-mobilenet ( / ) and https://github.com/Zehaos/MobileNet (66.51% / 87.09%) + paper
% Googlenet accuracy: https://github.com/albanie/convnet-burden
% Googlenet flops: https://github.com/albanie/convnet-burden
% Shufflenet accuracy: Learning Transferable Architectures for Scalable Image Recognition
% Squeezenet accuracy: SqueezeNet: AlexNet-level accuracy with 50x fewer parameters and <0.5MB model size

\setlength{\tabcolsep}{2pt}
\begin{minipage}{.6\textwidth}
\begin{table}[H]
\def\arraystretch{1.5}
\small
\begin{tabular}{lp{1.8cm}lll}
\bf Dataset 	& \bf Network & \bf Params & \bf FLOPS & \bf Top1/ Top5 (\%)
\\ \hline
\multirow{2}{*}{CIFAR10}      & ResNet101 	 & 42 M 	 & 2.5G 	& 93.8	\\
						      & IamNN 		 & 4.5 M  	& 1.1G \tiny (.7G - 2G)		& 94.6	\\
\hline
\multirow{2}{*}{CIFAR100}     & ResNet101 	 & 43 M 	& 2.5G		& 79.3	\\
						      & IamNN 		 & 4.6 M  	& 1.6G	\tiny (.7G - 2G)	& 77.8 	\\
 \hline
\multirow{2}{*}{ImageNet }	  & ResNet152  	 & 60 M   	& 11.5 G	& 77.0 / 93.3	\\
							  & ResNet18	 & 12 M   	& 1.8 G   	& 69.5 / 89.2  \\
	\tiny \hspace{2pt} Single crop  & IamNN 1 iter& 4.8 M 	& 0.9 G		& 60.8 / 83.2  \\
	\tiny \hspace{2pt} Single network      & \textbf{IamNN}      	 & \textbf{5 M} 		& \textbf{4 B \tiny (2.5G - 9G)}		& \textbf{69.5 / 89.0}	\\
                              & ShaResNet34  & 14 M		& 11 G		& 71.0 / 91.5 	\\
\cline{2-5}
							  & Googlenet     & 7 M	 	& 1.6 G		& 65.8 / 87.1 	\\
                              & MobileNet1    & 4.2 M   & 570 M		& 70.6 / 89.5 	\\
                              & ShuffleNet2x  & 5.6 M  & 524 M		& 70.9 / 89.8  	\\
                              & SqueezeNet    & 1.3 M    & 830 M		& 57.5 / 80.3	\\
                              
\end{tabular}
\caption{Number of parameters and classification accuracy on three benchmark datasets for our IamNN architecture and the corresponding ResNet architecture.}
\end{table}
\label{tbl:results}
\end{minipage}%
\hspace{15pt}
\begin{minipage}{.4\textwidth}
\begin{figure}[H]
\begin{tabular}{cc}
	\begin{subfigure}[l]{0.4\linewidth}
		\includegraphics[width=\linewidth]{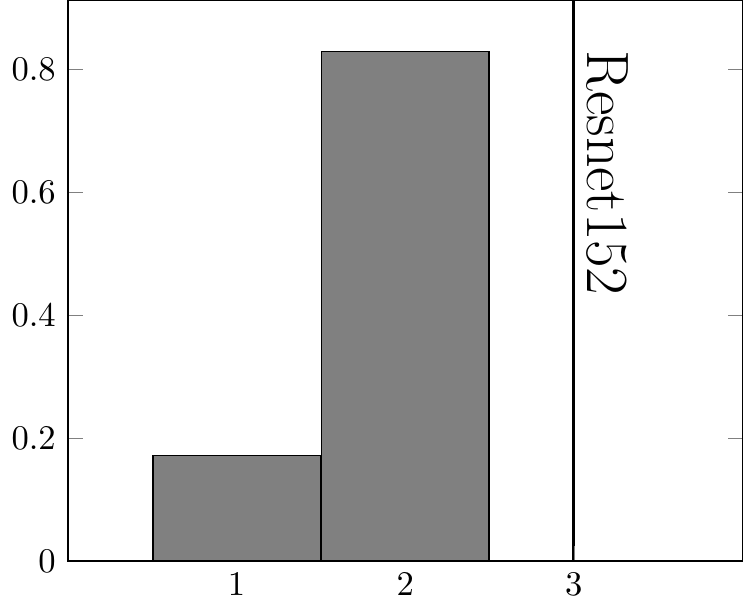}
    \end{subfigure}&
    \begin{subfigure}[l]{0.4\linewidth}
		\includegraphics[width=\linewidth]{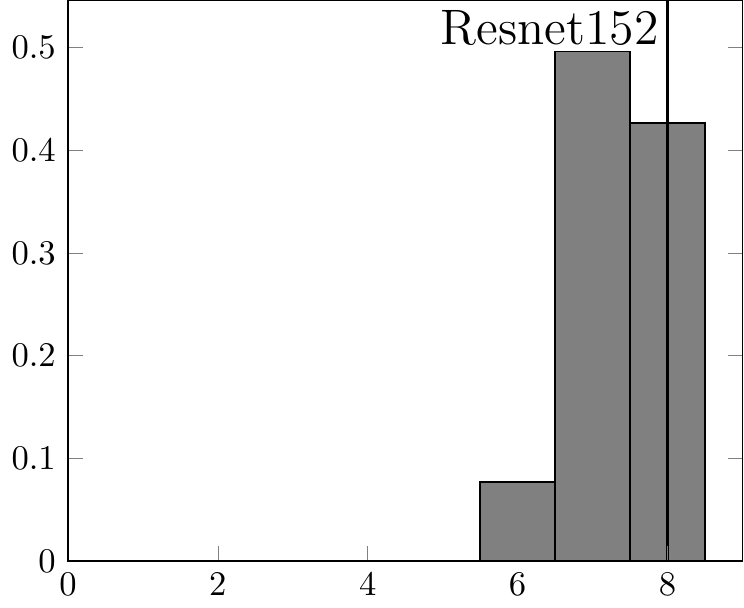}
    \end{subfigure}\\
    \begin{subfigure}[l]{0.4\linewidth}
		\includegraphics[width=\linewidth]{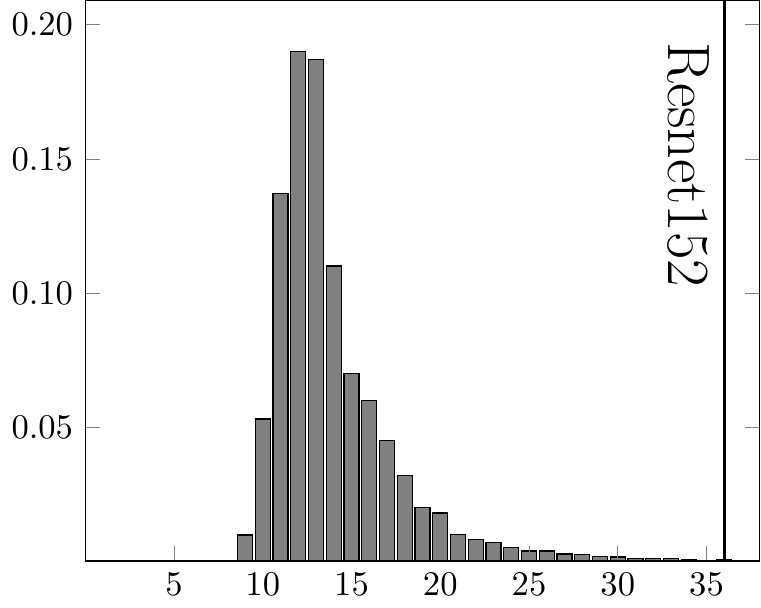}
    \end{subfigure}&
    \begin{subfigure}[l]{0.4\linewidth}
		\includegraphics[width=\linewidth]{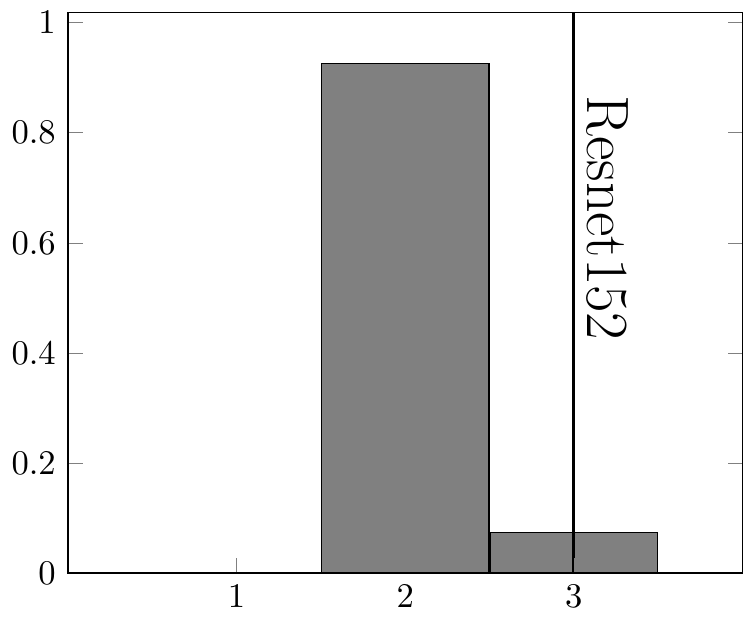}
    \end{subfigure}
\end{tabular}
\caption{Number of iterations used in each of the four blocks (ImageNet network)}
\label{fig:iterations}
\end{figure}
\end{minipage}%

\section{Conclusion and future work}
We proposed a new ResNet based architecture based on insights in the iterative refinement behavior of ResNets and Highway networks. By using the same set of weights multiple times we can reduce the model size by 90\%. Thanks to adaptive computation time the computation cost of the model depends on the complexity of the input image and is on average much lower than typical ResNets. In future work we will try to improve the results on the ImageNet dataset and we will incorporate other techniques such as depthwise separable convolutions to reduce the computational cost even further. 

\clearpage
%\textbf{\textcolor{red}{PM comments:}}
%\begin{itemize}
%\item We need to calculate ACT overhead and conclude if that is our bottleneck or not (in terms of GPU time ).
%\item results on flexibility of prediction with varying 
%$\tau$ are important.
%\item Need to show images of iterative feature map refinement
%\item Try to get activation maps for iterative estimation. 
%\item add top5
%\end{itemize}

\bibliography{iclr2018_workshop}
\bibliographystyle{iclr2018_workshop}

\end{document}